\documentclass[10.5pt,twocolumn]{article}
\usepackage{eai-original-template}
\usepackage{xurl}
\usepackage{microtype}
\usepackage{balance}
\usepackage{xcolor}
\usepackage{tikz}
\definecolor{revisionred}{RGB}{200,0,0}
\newcommand{\revision}[1]{#1}

\journalname{ENGINEERING Artificial Intelligence}
\articledoi{} % DOI to be assigned by the journal.
\articletype{PERSPECTIVE}
\papertitle{Autonomous Video Generation for Self-Evolving World Models}
\paperauthors{Xin Wang\textsuperscript{a,b}, Wenxuan Liu\textsuperscript{a,b}, Tongtong Feng\textsuperscript{a,b}, \correspondingauthormark{Wenwu Zhu\textsuperscript{a,b}}}
\addaffiliation{a}{Department of Computer Science and Technology, Tsinghua University, Haidian District, Beijing 100084, China}
\addaffiliation{b}{Beijing National Research Center for Information Science and Technology, Haidian District, Beijing 100084, China}
\copyrightstatement{\revision{\copyright\ The Author(s) 2026. This article is published by Higher Education Press.}}
\correspondingauthor{\revision{Wenwu Zhu, E-mail: wwzhu@tsinghua.edu.cn}}
\articlehistory{} % Editorial dates are not supplied; do not invent them.

\begin{document}

\begin{eaifrontmatter}
  \makeeaititle
  \begin{highlights}
    \item \revision{Autonomous video generation connects imagined futures to valid embodied action through counterfactual controllability.}
    \item \revision{Generation, Binding, Verification and Distillation form a feedback loop for self-evolving world models.}
    \item \revision{Novelty, consistency, robustness and efficiency provide complementary evaluation dimensions for action-valid imagination.}
  \end{highlights}
  \begin{abstract}
Large-scale video generation models are increasingly described as world models because they can learn rich spatiotemporal regularities from visual data.
We argue that an ideal world model should benefit in a self-evolving generative character. As such, a self-evolving world model can be realized via a generative model conditioned by \revision{various constraints}, whose future predictions are refined through feedback. However, traditional visually plausible predictions alone are not enough to establish whether an imagined future is physically actionable for a particular embodied agent, failing to provide informative feedback from environments for self-evolving improvement. To realize self-evolving world models, this article proposes the concept of autonomous video generation, which is evaluated through counterfactual controllability, i.e., the ability to i) generate intervention-conditioned futures, ii) bind these future frames to embodiment constraints, iii) verify them under distribution shifts, and iv) distil surviving branches into compact variables for decision-making. We formalize a four-stage closed-loop optimization of Generation, Binding, Verification and Distillation, together with four corresponding evaluation metrics: novelty, consistency, out-of-distribution (OOD) and efficiency. We further discuss two examples, i.e., drones and manipulators, as early embodied testbeds where wind, sensing limits, actuation delay, contact dynamics and recovery constraints can be systematically perturbed and verified. The central claim is that the framework of autonomous video generation for self-evolving world models should not be judged by video fidelity alone, but by whether the generated frames improve valid action under counterfactual interventions and various embodiment constraints.
  \end{abstract}
  \vspace{8pt}
\end{eaifrontmatter}

\section{Video Generation as A Partial World Model}
Existing literature \cite{sora2024world} claims that video generation essentially is world modelling. On the one hand, the claim is productive because it pushes generative AI beyond static images and toward temporally extended physical scenes. On the other hand, this claim dangerously relies on the belief that scaling visual prediction alone will automatically yield physical agents. 
We prefer a more accurate statement: video generation models learn a partial, implicit spatiotemporal world model, but not a fully grounded or controllable one.
The reason is as follows: a model may generate a plausible video of a drone crossing a forest or a robot arm manipulating a cup, yet still fail to know which variables are controllable, which constraints belong to a particular body and which futures remain valid under intervention. 

\begin{figure*}[t]
\centering
\includegraphics[width=0.98\textwidth]{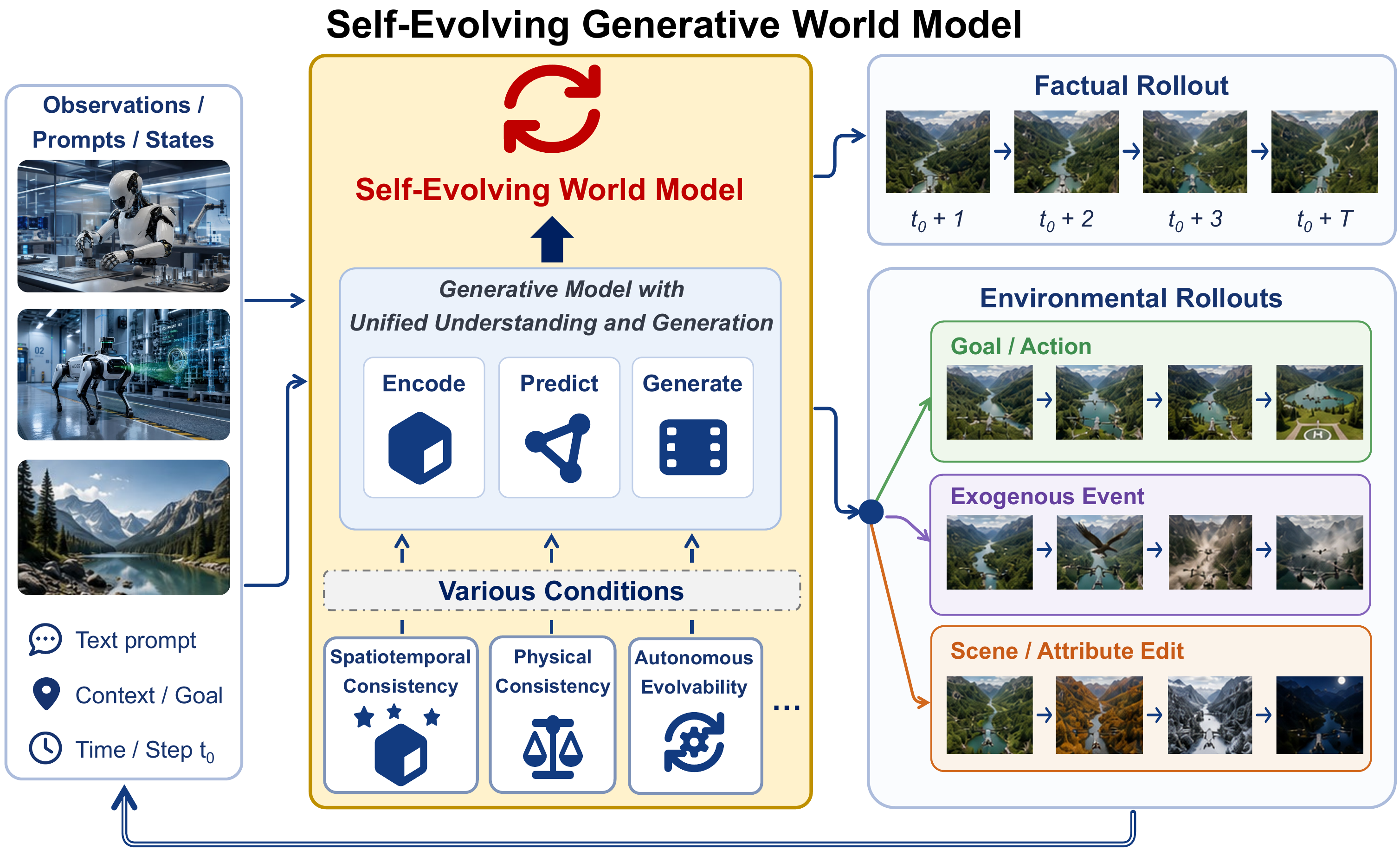}
\caption{The conceptual framework of \textit{Self-Evolving Generative World Model}, where initial observations, prompts and states are processed by a generative model that unifies understanding and generation, with various conditions such as spatiotemporal consistency, physical consistency and autonomous evolvability to constrain the generated futures. The generative model is able to produce a factual rollout and environmental rollouts representing controllable alternatives under goal/action changes, exogenous events and scene/attribute edits.}\label{fig:self-evolving generative WM}
\end{figure*}

The frontier in essence is not predictive realism alone, instead it emphasizes a self-evolving generative nature that requires the generative model \revision{to possess the ability} to iteratively evolve as well. Fig.~\ref{fig:self-evolving generative WM} demonstrates one conceptual framework of such \textit{Self-Evolving Generative World Model} where the core (video) generative model is capable of unifying understanding and generation, conditioned with constraints including spatiotemporal consistency, physical consistency, autonomous evolvability etc. This conceptual framework can be easily instantiated via the family of diffusion based generative models with various conditions added during the training phase to produce diverse rollouts.
Nevertheless, we argue that an ideal self-evolving generative \revision{world model requires} the decisive criterion to be counterfactual controllability: the capability of asking what would happen under an action, to test whether the generated future can survive embodiment constraints and to feed the resulting action knowledge back into future imagination (generation).

%The key challenge is therefore not predictive realism alone, but closing the loop between generation, evaluation and adaptation. In this sense, a self-evolving world model is a self-evolving generative model: counterfactual futures are tested for controllability, consistency and physical validity, and the resulting feedback reshapes subsequent generation.

%The frontier in essence is not predictive realism alone, instead it emphasizes a self-evolving generative nature that 

Therefore, in this paper we present a new perspective, i.e., autonomous video generation with counterfactual controllability is one promising way to realize self-evolving world models. %\revision{Figure~\ref{fig:self-evolving generative WM} summarizes this view.}

\section{Autonomous Video Generation: From Plausible Video to Valid Action}

A counterfactual embodied world model does not merely estimate future frames $p(o_{t+1:t+k}
\mid o_{\leq t})$, where $o_{\leq t}$ denotes the observation by time $t$ and $o_{t+1:t+k}$ denotes the observation during time $t+1$ to $t+k$. In this logic, the formal target is an intervention-conditioned future
$p(\tau\mid s_t,\operatorname{do}(a_{t:t+k}),e)$, where $k$ is the prediction horizon, $s_t$ is the latent state inferred from past observations and memory, $a_{t:t+k}$ is the candidate action sequence, $\operatorname{do}(\cdot)$ denotes an intervention rather than passive conditioning, $\tau$ is an imagined future trajectory and $e$ denotes embodiment constraints: the agent's body, sensors, actuation, controller and energy budget.

\revision{Here, we use counterfactual controllability prospectively: given the same context, interventions specify alternative actions, and their predicted outcomes are compared for embodied feasibility. Action-conditioned prediction thus supplies the futures used in this comparison.}
 
\begin{figure*}[t]
\centering
\includegraphics[width=0.98\textwidth]{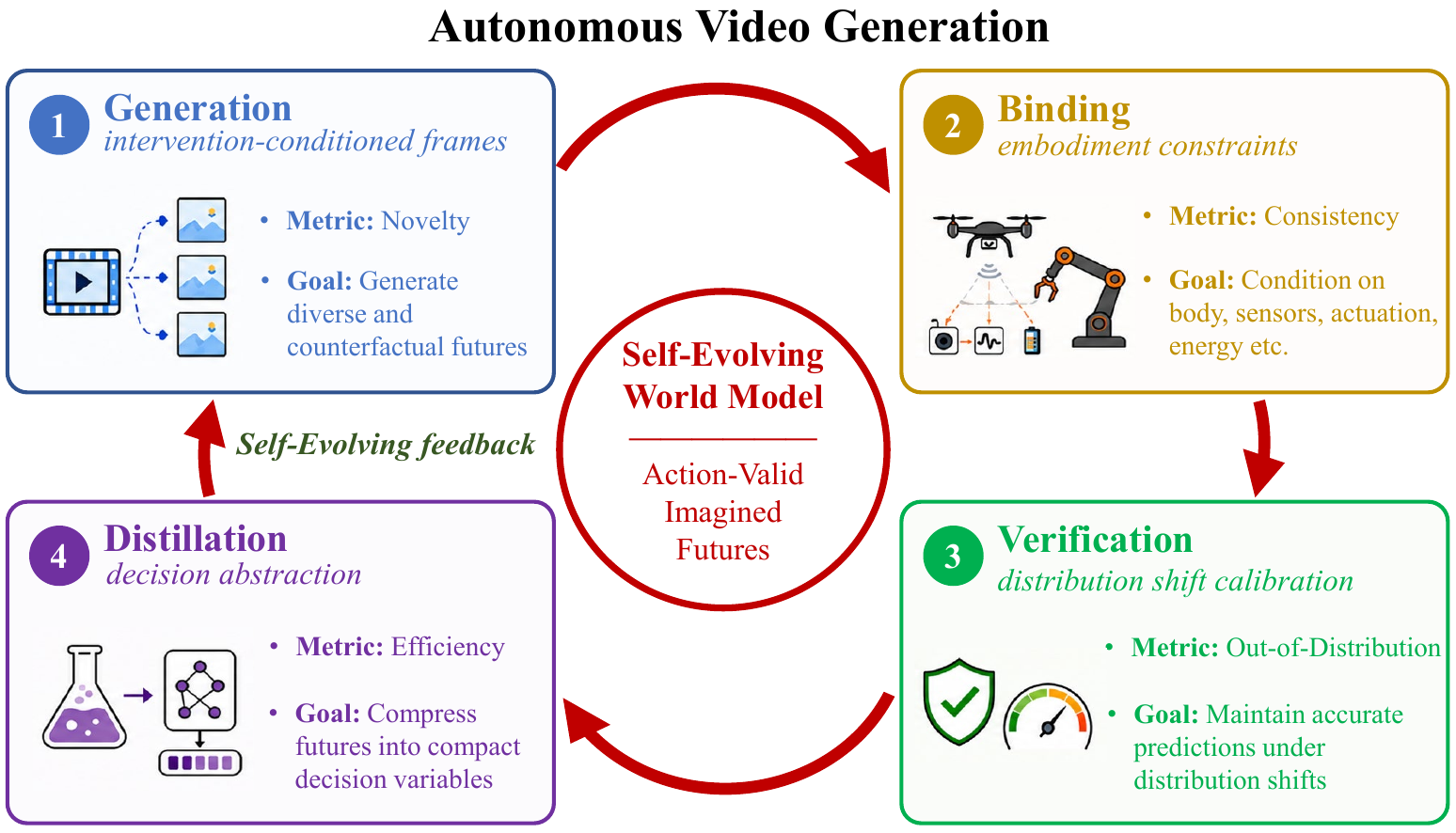}
\caption{Autonomous video generation with counterfactual controllability, where the framework is joint optimization of Generation, Binding, Verification and Distillation in a closed-loop manner. Generation proposes intervention-conditioned future frames and is evaluated by novelty; Binding attaches these future frames to embodiment constraints and is evaluated by consistency; Verification performs calibration and is evaluated by Out-of-Distribution; Distillation conducts decision abstraction and is evaluated by efficiency. The self-evolving feedback from Distillation to Generation completes the function of a self-evolving world model with action-valid imagined future.}\label{fig:gbvd}
\end{figure*}

We define counterfactual controllability as the capacity of a self-evolving world model to produce action-valid imagined futures through four closed-loop stages: 1) \textbf{Generation} proposes diverse counterfactual future frames, 2) \textbf{Binding} conditions these frames on embodiment constraints, 3) \textbf{Verification} discovers and calibrates drifted branches under shift, and 4) \textbf{Distillation} compresses surviving branches into compact decision variables. The four stages are connected with each other in a closed-loop manner, supporting the establishment of a self-evolving world model, as shown in Fig.~\ref{fig:gbvd}.

\revision{Verification feedback can come from simulation, physical execution or explicit constraint checks, supporting branch rejection and prediction calibration.}
 %are normalized operational axes rather than universal scalar laws. 

They extend traditional generative AI paradigm toward an embodied and intervention-conditioned setting, where frame likelihood, sample fidelity, distributional coverage, and decision cost may capture different properties and should be seriously considered by video generation models with respect to a particular emphasis. \cite{theis2016evaluation,kynkaanniemi2019precision}.
%judged with respect to the downstream applications \cite{theis2016evaluation,kynkaanniemi2019precision}. 
Autonomous video generation, with an emphasis on \textit{valid action}, jointly optimizes the closed-loop stages during video generations with four corresponding metrics: i) Novelty evaluates Generation to promote diversity, ii) Consistency evaluates Binding to promote embodiment, iii) Out-of-Distribution evaluates Verification to promote generalization and iv) Efficiency evaluates Distillation to promote feasibility. 
We will elaborate each evaluation metric with quantitative definitions in detail.

%These metrics correspond to the four failures that the loop is designed to resolve: generation without novelty, embodiment mismatch, brittle predictions under shift and decision overload without abstraction.

%The four axes in Figure~\ref{fig:gbvd} can be treated as compact evaluation lenses that organize familiar quantities from video generation, OOD evaluation and embodied control. In concrete benchmarks, each \(M_j\) can be reported on a high-is-better scale before aggregation.

\textit{Novelty.}
Novelty should not be confused with unconstrained diversity. A useful generator must not only produce futures that are diverse enough to expand the counterfactual space, but still plausible enough to remain useful for subsequent binding and verification as well. We define:
\begin{equation}
M_{\mathrm{nov}}
=
\frac{
2D_{\mathrm{div}}Q_{\mathrm{vid}}
}{
D_{\mathrm{div}}+Q_{\mathrm{vid}}
}.
\label{eq:nov}
\end{equation}
Here \(D_{\mathrm{div}}\) denotes an LPIPS-based perceptual diversity score, and \(Q_{\mathrm{vid}}\) denotes an FVD-based high-is-better video fidelity score \cite{zhang2018lpips,unterthiner2018fvd}. The harmonic form is deliberate: high diversity with poor video fidelity suggests uncontrolled divergence, whereas high fidelity with low diversity suggests conservative prediction. As such, \textbf{Generation} will be ``ideal'' only when the imagined futures are sufficiently diverse while retaining video-level plausibility.

\textit{Consistency.}
Consistency examines whether \textbf{Binding} ensures the frames as a valid video, a physical trajectory and an executable future for the embodied agent. We define:
\begin{equation}
M_{\mathrm{con}}
=
\left(
C_{\mathrm{vid}}
C_{\mathrm{phy}}
C_{\mathrm{emb}}
\right)^{1/3}.
\label{eq:con}
\end{equation}
Here \(C_{\mathrm{vid}}\) denotes video-level consistency, which can be instantiated by temporal flickering, motion smoothness or related video-generation benchmark dimensions; \(C_{\mathrm{phy}}\) denotes physical plausibility of motion and interaction; and \(C_{\mathrm{emb}}\) denotes embodiment feasibility under body, sensing, actuation, control and energy constraints \cite{huang2024vbench,sanchezgonzalez2020physics,chi2025diffusion}. The geometric form emphasizes that video coherence, physical feasibility and embodiment feasibility are jointly required for a generated future to be useful for control.

\textit{Out-of-Distribution (OOD).}
OOD evaluation resorts to the standard OOD benchmark retention setting. Let \(\mathcal D_{\mathrm{ID}}\) be the in-distribution (ID) benchmark and \(\{\mathcal D_{\mathrm{OOD}}^{(r)}\}_{r=1}^{R}\) be the set of OOD benchmark variants, e.g., corrupted, perturbed or shifted test sets. We define:
\begin{equation}
M_{\mathrm{rob}}
=
\frac{1}{R}
\sum_{r=1}^{R}
\frac{
A(\theta;\mathcal D_{\mathrm{OOD}}^{(r)})
}{
A(\theta;\mathcal D_{\mathrm{ID}})
}.
\label{eq:rob}
\end{equation}
Here \(A(\theta;\mathcal D)\) is task success, prediction accuracy (ID scenario) or calibrated prediction accuracy (OOD scenario) of model \(\theta\) on benchmark \(\mathcal D\). This follows the common practice: report how much performance is retained from an ID benchmark to corrupted, perturbed or in-the-wild OOD benchmarks \cite{hendrycks2019robustness,koh2021wilds,ovadia2019uncertainty}.

\textit{Efficiency.}
Efficiency asks whether \textbf{Distillation} reduces decision overload. Let \(\mathcal B\) denote the set of budget types, such as planning time, computation, memory and retained decision variables etc. For each budget type \(b \in \mathcal B\), let \(S_b(r)\) be the task success rate (or the normalized score) given a particular available resource \(r\). We define:
\begin{equation}
M_{\mathrm{eff}}
=
\sum_{b\in\mathcal B}
\alpha_b
\int S_b(r)\,dr .
\label{eq:eff}
\end{equation}
Here \(\alpha_b\) weights the importance of budget type \(b \). The expression summarizes budget-performance AUCs across resource dimensions: an efficient distillation allows compact decision variables to support task success with less planning time, computation, memory or representational burden.

The four metrics can then be jointly optimized via a counterfactual-controllability score:
\begin{equation}
\mathcal M
=
\prod_{j\in\mathcal J} M_j^{w_j}.
\label{eq:mcc}
\end{equation}
Here \(\mathcal J=\{\mathrm{nov},\mathrm{con},\mathrm{rob},\mathrm{eff}\}\), and the non-negative weights \(w_j\) sum to 1. The product form treats action validity as a conjunction of four necessary stages. A future must add meaningful counterfactual variation, remain consistent with the world and the body, survive distribution shift and be compressed into efficient decision variables. If any stage fails, the overall controllability score will be strongly reduced. Equivalently,
\begin{equation}
\log \mathcal M
=
\sum_{j\in\mathcal J}
w_j\log M_j,
\qquad
\frac{\partial\log \mathcal M}{\partial\log M_j}
=
w_j .
\label{eq:mcc-sensitivity}
\end{equation}
Thus \(w_j\) is the log-scale sensitivity of counterfactual controllability to metric \(j\). This gives the loop a compact mathematical interpretation: Novelty, Consistency, Out-of-Distribution and Efficiency together act as coupled requirements that jointly determine whether an imagined future is actionable.

\revision{The four dimensions assess alternative coverage, embodied feasibility, reliability under shift and decision efficiency, respectively. A high combined score summarizes these complementary qualities under the evaluation protocol, without guaranteeing validity for every intervention. Applications may emphasize different dimensions: robustness and efficiency for drones, or physical consistency for contact-rich manipulation.}

\textbf{Summary.} The above definitions also clarify the role of each stage in making those generated future frames accountable to embodied action.
\begin{itemize}
    \item \textbf{Generation} discourages redundancy. A video model may synthesize many futures while merely recycling similar trajectories, whereas useful intervention-conditioned futures should be both diverse and video-plausible. Perceptual diversity and video-fidelity metrics help separate useful variation from uncontrolled divergence, and interactive simulators similarly point beyond static visual realism toward action-conditioned outcomes \cite{zhang2018lpips,unterthiner2018fvd,yang2024unisim}.
    
    \item \textbf{Binding} prevents embodiment mismatch. A scene may respond to prompts while ignoring body morphology, sensors, actuation limits, energy constraints, rotor dynamics, actuator delay, gripper force limits or sensor blind spots. Video-level benchmarks, learned physical simulation and visuomotor diffusion policies show why temporal consistency, physical interaction and executable action distributions must be explicitly represented rather than implicitly inferred from appearance alone \cite{huang2024vbench,sanchezgonzalez2020physics,chi2025diffusion}.
    
    \item \textbf{Verification} addresses prediction failures under distribution shift. An imagined branch of frames may look plausible at the video level while become infeasible or even unsafe under changing observations, dynamics or embodiment conditions. Verification exposes out-of-distribution failures and calibrates whether each imagined branch should be accepted, rejected or down-weighted \cite{hendrycks2019robustness,koh2021wilds,ovadia2019uncertainty}.
    
    \item \textbf{Distillation} prohibits decision overload without abstraction. Robots need compact decision variables rather than cinematic continuity for collision risk, contact mode, recovery basin, information gain and task value. Latent world-model agents demonstrate that imagined futures are useful when being compressed into representations that can improve controllability \cite{ha2018world,hafner2019planet,hafner2025dreamer}.
\end{itemize}

\revision{Video complements latent- or state-space prediction when visually rich changes, such as occlusion or deformation, matter for decisions; Distillation extracts their compact control-relevant consequences.}

%The goal is not simply to make videos autonomous, but to make video imagination accountable to embodied action.

\section{Scaling Is Not Enough}
Recent progress not only makes generative world models increasingly important for embodied AI, but also shows that scale alone is not sufficient. Large video models indicate that generative prediction can capture rich visual regularities of the physical world, offering a scalable prior over possible futures rather than a simulator by itself \cite{sora2024world}. Model-based reinforcement learning has long demonstrated the value of imagined dynamics for control, from recurrent world models \cite{ha2018world} and latent planning from pixels \cite{hafner2019planet} to Dreamer-style agents that learn policies through latent future rollouts \cite{hafner2025dreamer}. Robot foundation models such as RT-2 \cite{zitkovich2023rt2}, $\pi_0$ \cite{black2024pi0}, VIMA \cite{jiang2023vima} and Open X-Embodiment \cite{oneill2023openx} show that large-scale semantic and robot data can support action generation, yet their behaviors remain tied to the coverage of demonstrations, tasks and embodiments. Generative policies, interactive environments and self-supervised video models further suggest that generative modelling can represent multimodal actions, controllable scenes and physical representations \cite{bruce2024genie,yang2024unisim,agarwal2025cosmos,assran2025vjepa2}.

Nevertheless, these lines remain only partially coupled. World-model agents imagine, but often optimize reward without explicitly rejecting physically invalid branches. Robot foundation models act, but are still bounded by demonstrated experience and embodiment-specific data. Generative simulators create worlds, but do not automatically prove that a specific body can exploit them safely. This is the causal gap: predicting what may happen in observed data is different from estimating what will happen under a specific action, embodiment and environment shift \cite{pearl2009causality,causalworld2024}. The missing component is therefore an embodied criterion for generative futures: which branches are physically actionable, which should be rejected under distribution shift, and which can be compressed into decision variables for control. This requires checking feasibility under embodiment constraints, environmental variation, and uncertainty in the learned dynamics.

\section{Embodied Testbeds Are Necessary}

Drones and manipulators provide useful early benchmark beds for self-evolving world models, not because they are more important than humanoids, but because their constraints can be measured, perturbed and verified. A drone world model that only predicts visual motion is insufficient. For instance, a self-evolving drone world model must i) generate intervention-conditioned futures under side wind, payload shift, visual aliasing, battery limits and delayed control; ii) bind these futures to aerodynamics, sensing, actuation and energy constraints; iii) verify these generated futures under distribution shift through falsification and calibration; iv) distil the accepted branches into compact variables for observability, energy consistency, risk and recovery. The decisive cases are rare in demonstrations but critical in deployment: gust-induced drift, sensor dropout, rotor degradation and recovery from poor approach geometry.

Manipulation provides a complementary test bed because its most difficult counterfactuals arise through contact. If the gripper nudges an object before grasping, will it roll, deform, jam or reveal a better affordance? If the object slips, can the policy recover, or has an early action made the goal unreachable? Such questions are not answered by video realism alone. They require \textbf{Generation} to propose diverse contact futures, \textbf{Binding} to condition them on body, sensing, actuation, force and reachability constraints, \textbf{Verification} to reject infeasible or unsafe branches under contact and perception shifts, and \textbf{Distillation} to compress long-horizon consequences into decision variables for grasp stability, recoverability and task value. Systems that pass these tests would not merely animate plausible manipulation videos, but identify which imagined contacts can become controllable affordances as well.

%\section*{The self-evolving loop}

\section{Concluding Remarks}

\textit{The self-evolving loop.} 
%Figure~\ref{fig:gbvd} illustrates the closed loop. 
\textbf{Generation} produces otherwise inaccessible intervention-conditioned futures: gusts, failed grasps, occlusions, deformable contacts, near collisions and partial task failures. Its metric is novelty: whether the model proposes diverse counterfactual futures beyond familiar rollouts while retaining video-level plausibility \cite{zhang2018lpips,unterthiner2018fvd}. \textbf{Binding} attaches these futures to embodiment constraints by conditioning on body, sensors, actuation and energy, including aerodynamics, reachability, force closure, latency, sensing limits and controller bandwidth. Its metric is consistency with video-level coherence, physical interaction and executable action constraints \cite{huang2024vbench,sanchezgonzalez2020physics,chi2025diffusion}. %Verify is not a perfect oracle. It measures robustness through falsification and calibration: analytic tests, learned consistency checks and sim-to-real calibration, including safety filters such as control barrier functions when intervention risk must be bounded \cite{hendrycks2019robustness,koh2021wilds,ovadia2019uncertainty,ames2019cbf}. 
\textbf{Verification} stress-tests bound futures under distribution shift by perturbing observations, dynamics and embodiment conditions, including sensor noise, actuation delay, wind changes or unseen objects. Its metric is Out-of-Distribution: whether imagined branches remain calibrated, action-valid and safe beyond the training distribution, with infeasible or unsafe futures rejected or down-weighted \cite{hendrycks2019robustness,koh2021wilds,ovadia2019uncertainty,ames2019cbf}.
\textbf{Distillation} performs decision abstraction. It measures efficiency by compressing accepted branches into compact decision variables: a map from imagined futures to risk, reachability, generalization and task value, in line with latent world models that use imagined futures to support downstream control \cite{hafner2025dreamer}.

The loop becomes self-evolving through the arrow pointing from \textbf{Distillation} back to \textbf{Generation}. Through the self-evolving feedback, verified action knowledge progressively reshapes the model's generative distribution. This self-evolving feedback is not unconstrained self-training on synthetic videos, instead it is the return of distilled action knowledge to the next round of imagination (generation). Rejected futures can suppress inconsistent branches; high-novelty branches can become new generation priors; brittle regimes can trigger targeted falsification and calibration; and efficient abstractions can bias future imagination toward compact, controllable plans. In this sense, autonomous video generation is not a one-way pipeline from prompt to clip, instead it serves as an iterative mechanism for improving what should be imagined, what should be tested and what should be carried forward into action-valid control.

\revision{Accumulated experience can refine generation priors and guide future verification. Prediction informs action, while observed action outcomes improve subsequent predictions, sustaining long-term self-evolution.}

\textit{Evaluation beyond video fidelity.}
Generative world models require evaluation beyond whether generated videos look realistic. Video loss, visual preference and prompt adherence measure observational fidelity, but not whether an imagined future is actionable for a particular embodiment. Embodied benchmarks should instead test counterfactual utility: whether imagined branches improve closed-loop decisions under controlled interventions, distribution shift and safety constraints. Take drones as an instance, test suites should perturb wind, GPS availability, visual aliasing, payload, battery state and actuator health, and measure recovery, constraint violation, energy consistency and calibrated risk. Take manipulators as another example, they should vary hidden mass, clutter, deformability, occlusion and tool availability, and measure grasp stability, contact feasibility, replanning and recoverability. A strong model should not merely generate plausible futures; it should reject infeasible branches, revise plans under shift and expose useful affordances, while accounting for embodiment constraints and uncertainty throughout task execution.

The resulting criterion is counterfactual controllability. A self-evolving world model should generate intervention-conditioned futures with novelty, bind them to embodiment constraints with consistency, verify them under distribution shift for generalizability, and distil them into compact decision variables for efficient control. The overall value is measured not by video fidelity alone, but by whether imagination improves valid actions. This requires explicit tests of physical feasibility, uncertainty calibration, and improvements in task performance under controlled changes in dynamics.

\section*{CRediT authorship contribution statement}
% Please verify and revise the following CRediT roles before submission.
Xin Wang: Conceptualization, Methodology, Writing--original draft. Wenxuan Liu: Methodology, Writing--review and editing. Tongtong Feng: Visualization, Writing--review and editing. Wenwu Zhu: Supervision, Project administration, Funding acquisition, Writing--review and editing.

\section*{Acknowledgments}
Funding: This work was supported by Beijing National Research Center for Information Science and Technology [grant number BNR2026TD03005]. The funding source had no role in the preparation of the manuscript or the decision to submit it for publication.

\section*{\revision{Competing interests}}
\revision{The authors declare no conflicts of interest.}

\section*{Declaration of generative AI and AI-assisted technologies in the manuscript preparation process}
During the preparation of this work, the authors used ChatGPT (OpenAI) to check typos for the Engineering submission. After using this tool, the authors reviewed and edited the content as needed and take full responsibility for the content of the published article.

\section*{Data availability}
No new data were generated or analyzed in this work.

\end{document}